\documentclass{article}

\usepackage[numbers]{natbib}
\usepackage[final]{nips_2017}
\usepackage[utf8]{inputenc} % allow utf-8 input
\usepackage[T1]{fontenc}    % use 8-bit T1 fonts
\usepackage[pdftex,colorlinks]{hyperref}       % hyperlinks
\usepackage{url}            % simple URL typesetting
\usepackage{booktabs}       % professional-quality tables
\usepackage{amsfonts}       % blackboard math symbols
\usepackage{nicefrac}       % compact symbols for 1/2, etc.
\usepackage{microtype}      % microtypography
\usepackage{color}
\usepackage{graphicx}
\usepackage{subfigure}
\usepackage{caption}
\usepackage{amsmath}
\usepackage{multirow}
\usepackage{amssymb}
\usepackage{bm}
\usepackage{caption}
\usepackage{wrapfig}
\usepackage{floatrow}
\usepackage{amsthm}
\newfloatcommand{capbtabbox}{table}[][\FBwidth]
\newtheorem{theorem}{Theorem}

\title{Rethinking Feature Discrimination  and        Polymerization for Large-scale  Recognition}

\author{
  Yu Liu\thanks{The first two authors contribute equally.}$^{~}$, Hongyang Li$^{*}$, Xiaogang Wang
    \smallskip
  \\
  The Chinese University of Hong Kong\\
  \texttt{\{yuliu, yangli, xgwang\}@ee.cuhk.edu.hk} \\
}

\begin{document}
% \nipsfinalcopy is no longer used

\maketitle

\begin{abstract}
	Feature matters. 
	%\\
	%\fbox{\rule{0pt}{2.4in} \rule{0.9\linewidth}{0pt}}
	How to train a deep  network to acquire discriminative features across categories and polymerized features within classes has always been at the core of many computer vision tasks, specially for large-scale recognition systems where  test identities are unseen during training and the number of  classes could be at million scale. In this paper, we address this problem based on the simple intuition that the cosine distance of features in high-dimensional space should be close enough within one class and far away across categories. To this end, we proposed the congenerous cosine (COCO)  algorithm to simultaneously optimize the cosine similarity among data. It inherits the softmax property to make inter-class features discriminative as well as shares the idea of class centroid in metric learning. Unlike previous work where the center is a temporal, statistical variable within one mini-batch during training, the formulated centroid is responsible for clustering inner-class features to enforce them polymerized around the network truncus. COCO is bundled with discriminative training and learned end-to-end with stable convergence. Experiments on five benchmarks have been extensively conducted to verify the effectiveness of our approach on both small-scale classification task and large-scale human recognition problem.
\end{abstract}

\section{Introduction}\label{sec:intro}

Deep neural networks \cite{alexnet,resNet,hinton_nature} have dramatically advanced the computer vision community in recent years, 
with high performance boost in tremendous tasks, such as image classification
%TODO: use your own citation after publication.
\cite{,
	resNet,vgg,li2016multi}, object detection \cite{fast_rcnn,faster_rcnn,li2017zoom,wanli_iccv,liu2017recurrent}, 
object tracking \cite{czz_tracking}, saliency detection \cite{li2016cnn_sal}, 
\textit{etc}.
The essence behind the success of deep learning 
resides in both the superior expression power of non-linear complexity in high-dimension space \cite{hinton_nature}
and the large-scale datasets \cite{imagenet_conf,celebrity} where the deep networks could, in full extent, learn complicated patterns and representative features. The basic question still halts the core of many vision tasks: \textit{how can we obtain discriminative features across categories and polymerized features within one class?}
%In this work, we extend %pursuit 
%the ability of deep models to a higher level in order to make features of 
%the human recognition task more robust and generalized, based on the state-of-the-art detector \cite{faster_rcnn} to recognize face and body,
%and  pose estimator \cite{pose} to predict key points of an identity.

The most fundamental task in computer vision is image classification \cite{vgg,alexnet,resNet}: given $N$ categories in the training set, we train a deep model that could best discriminate features among categories in high-dimensional space. The test set also consists of the same $N$ classes (identities). However, there are two challenges when we apply the classification task in real-world applications. The first is the number of categories could be exponentially increased (\textit{e.g.}, MegaFace \cite{kemelmacher2016megaface} contains one million test identities). This requires the features should be \textit{discriminative} to have large margin across classes. The second is the category mismatch between training and test set. Often we train on a set of $N$ known classes and evaluate the algorithm on $M$ new, different identities.
This hastens the features should be \textit{polymerized} to share small distances within classes.

In this paper, we %try to
 optimize the deeply learned features to be both discriminative and polymerized, given the large-scale constraint: ultra category number and open test set.
Faced with the aforementioned challenges, the community extends the classification problem into two sub tasks: data verification and identification. Since our main focus lies in the field of human-related tasks, we use the term face with data exchangeably thereafter.
Face verification \cite{feng_wang_mm,l2_constrained} aims at differentiating whether a pair of instances belong to the same identity; whilst face identification \cite{sphereFace,contrastive_loss} strives for predicting the class of an identity given the test gallery \cite{kemelmacher2016megaface}. Person recognition \cite{oh_iccv,piper} also belongs to the verification case, only with more non-frontal and unstrained face (head) settings\footnote{ The term recognition has a broad sense: detection, verification, classification, \textit{etc}. Person recognition is face verification with more unstrained samples. For not making any confusion, we follow this terminology as did in the community. In this paper, we refer to face verification, identification and person recognition, in a general sense, as \textit{human recognition}.}.

\begin{figure}[t]
	\centering
	%\fbox{\rule{0pt}{2in} \rule{0.9\linewidth}{0pt}}
	\includegraphics[width=.97\textwidth]{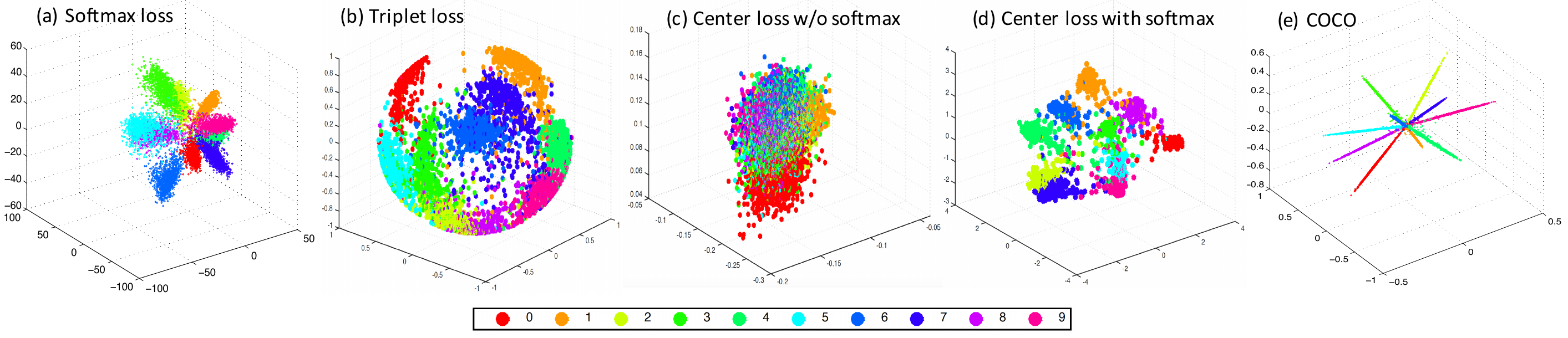}
	%\vspace{-.1cm}
	\caption{Feature visualization under different loss strategies, trained on MNIST \cite{mnist} with 10 classes. 
		%The feature dimension is three.
		(a) Softmax loss; (b) triplet loss; (c) center loss without softmax; (d) center loss; (e) our proposed COCO.
		The feature dimension is three for better illustration. The triplet loss \cite{facenet} has a constraint on feature embedding ($\| \bm{f} \|_2=1$) and thus the learned features are scatted on a hypersphere.
	}
	\label{fig:loss_motivation}
\end{figure}

For the feature discrimination concern, the softmax loss and its variants \cite{quite_similar_loss_to_coco,l2_constrained,sphereFace} are proposed to provide a hyperplane with large margin in high-dimensional space and thus effectively distinguish features across classes. However, it ignores the intra-class similarity, making the sample distance within one class distant also and thus amply filling in the feature space (Fig. \ref{fig:loss_motivation}(a)); for the feature polymerization concern, one can resort to distance metric learning \cite{eric_xing,metric_learning2}. Previous work \cite{face_nips16} usually introduced neural networks, using contrastive loss \cite{contrastive_loss} or triplet loss \cite{triplet_loss_xiaolong,loss,facenet}, to learn the features, followed by a simple metric such as Euclidean or cosine distance to verify the identity (Fig. \ref{fig:loss_motivation}(b)). Such methods are effective to decrease the intra-class distance in some extent and yet bears an internal drawback due to the formulation: the selection of positive and negative samples are relative within each batch and thus could occur training instability, especially the number of classes and data scale increase (see Fig. \ref{fig:loss_temp}(b)).

An alternative is to jointly combine the discriminative feature learning in a softmax spirit and the polymerized feature learning in a metric learning manner. The center loss \cite{center_loss} and its variant \cite{range_loss} are two typical approaches. They linearly combine two terms as the total loss: the first being softmax and the second a class center supervision $\| \bm{x}_i - \bm{c}_{l_i}\|$. The hybrid solution can effectively enlarge the inter-class distance as well as the inner-class similarity; nevertheless, the concept of class center is updated in a statistical manner without consulting to the network parameter update  during training. Moreover, the center loss term has to be bundled with an ad-hoc softmax (thus requires more model parameters); if not, the center barely changes according to its formulation and thus the loss could be zero (\textit{c.f.}, Fig. \ref{fig:loss_motivation}(c)-(d)).

To this end, we propose the congenerous cosine (COCO) algorithm to provide a new perspective on jointly considering the feature discrimination and polymerization.
The intuition is that we directly optimize and compare the cosine distance (similarity) between  features: the intra-class distance should be as minimal as possible whereas the inter-class variation should be magnified to the most extent (see Fig. \ref{fig:loss_motivation}(e)). COCO inherits the softmax property to make features discriminative in the high-dimensional space as well as keeps the idea of class centroid. Note that unlike the statistical role of $ \bm{c}_{l_i}$ in \cite{center_loss}, the update of our centroid is performed simultaneously during the network training; and there is no additional loss term in the formulation. 
By virtue of class centroid bundled with discriminative training, COCO is learned end-to-end with stable convergence.
Experiments on two small-scale image classification datasets first verify the feasibility of our proposed approach and three large-scale human recognition benchmarks further demonstrate the effectiveness of applying COCO in an ultra class number and open test set environment. The source code is publicly available\footnote{\href{https://github.com/sciencefans/coco_loss}{\texttt{ https://github.com/sciencefans/coco\_loss }}}.

\section{Related Work}
The related work of different loss methods to optimize features is compared and discussed throughout the paper. In this section, we mainly focus on the face and person recognition applications.

\textbf{Face verification and identification} have been extensively investigated \cite{kemelmacher2016megaface,l2_constrained,sphereFace,face_nips16,contrastive_loss,center_loss} recently due to the dramatic demand in real-life applications.
Schroff \textit{et al.}  \cite{facenet} directly learned a mapping from face images to a compact Euclidean space where distances directly correspond to a measure of face similarity. 
 DeepID2  \cite{contrastive_loss} employed a contrastive loss with both the identification and verification supervision. It increased the dimension of hidden representations and added supervision to early convolutional layers.
Tadmor \textit{et al.} \cite{face_nips16} introduced a multibatch method to generate invariant face signatures through training pairs. The training tended to be faster to convergence by virtue of smaller variance of the estimator.
SphereFace \cite{sphereFace} formulated an angular softmax loss that enabled CNN to learn  discriminative features by imposing constraints on a hypersphere manifold.

% new related work put aside after coco
%\vspace{-.3cm}
\textbf{Person recognition} in photo albums \cite{original,oh_eccv,oh_iccv,person_recog_rnn,piper,zlin} aims at recognizing the identity of people in daily life photos, where such scenarios can be complex with cluttered background.
%
%Anguelov \textit{et al.} 
\cite{original} first address the problem by proposing a Markov random filed framework to combine all contextual cues to recognize the identity of persons.
Recently, Zhang \textit{et al.} 
\cite{piper} introduce a large-scale dataset called PIPA for this task. The test set on PIPA
is split into two subsets, namely \texttt{test\_0} and \texttt{test\_1} with roughly the same number of instances.
%, where over 60000 instances of around 2000 individuals are annotated. 
%\cite{piper}  accumulate the cues of poselet-level person recognizer trained by a deep model to compensate for pose variations.
%
%
%A complimentary multi-level contextual model is formulated in \cite{zlin} under the conditional random filed framework.
%
%
In \cite{oh_iccv}, a detailed analysis of different cues %, including body, head, attributes, scenes, 
is explicitly investigated and three additional test  splits are proposed for evaluation. 
%Li \textit{et al.} 
\cite{person_recog_rnn} embed scene and relation contexts in LSTM and formulate person recognition as a sequence prediction task.
Note that previous work \cite{piper,oh_iccv,zlin,person_recog_rnn} use the training set {only} for extracting features and a follow-up classifier (SVM or neural network) is \textit{trained} on  \texttt{test\_0}. The recognition system is evaluated on the \texttt{test\_1} set. We argue that 
such a practice is infeasible and ad hoc in realistic application since the second training on  \texttt{test\_0}   is auxiliary and needs re-training if new samples are added.

\section{Optimizing Congenerous Cosine Distance}

\subsection{COCO Formulation and Optimization}
Let $\bm{f}^{(i)}\in \mathbb{R}^{D}$ denote the feature vector of the $i$-th sample,% from region $r$, 
where $D$ is the feature dimension. %For brevity, we drop the superscript $r$ since each region model undergoes the same COCO training.
We first introduce the {cosine similarity} 
%$\mathcal{C}(\bm{f}^{(i)}, \bm{f}^{(j)})$ 
of two features %$\bm{f}^{(i)}, \bm{f}^{(j)} \in \mathbb{R}^D$ 
%from a mini-batch $\mathcal{B}$ 
%($i,j$ denotes the sample index within the batch, $D$ is the feature dimension) 
as:
\begin{equation}
\mathcal{C}({\bm{f}^{(i)}}, \bm{f}^{(j)}) =  %\frac
{ \bm{f}^{(i)T}  \cdot 
	\bm{f}^{(j)}}  \big / {   \|\bm{f}^{(i)} \|  %\cdot 
	\| \bm{f}^{(j)}\| }.
\end{equation}
%$
%where $\odot$ denotes the inner product of two vectors and each feature vector $\bm{f}\in \mathbb{R}^D$ is of $D$ dimension. 
The cosine similarity metric quantifies  how close two samples are in the feature space. A natural intuition to  a desirable loss is to increase the similarity of samples within a category and enlarge the centroid distance of samples across classes. Let $l_{i}, l_{j} \in \{1, \cdots, K\}$ be the labels of sample $i, j$, where $K$ is the total number of categories,
% and $\bm{W}$ be the weights of a network, 
we have the 
%object function 
following loss
%as follows 
to \textit{maximize}:
%\begin{align}
%%& \arg \max_{\bm{W}^*} L(\mathcal{B}, \bm{W}), &\\
% &L(\mathcal{B}, \bm{W}) = \nonumber \\
% & \sum_{i,j \in \mathcal{B}} \delta(l_{i}, l_{j}) \mathcal{C}(\bm{{f}}^{(i)}, \bm{{f}}^{(j)}) -
% \big[1 - \delta(l_{i}, l_{j}) \big] \mathcal{C}(\bm{{f}}^{(i)}, \bm{{f}}^{(j)}), \label{basic_loss}
%\end{align}
\begin{gather}
\mathcal{L}^{naive} = \sum_{i,j \in \mathcal{B}} \frac{\delta(l_{i}, l_{j}) \mathcal{C}(\bm{{f}}^{(i)}, \bm{{f}}^{(j)}) }
{[ 1 - \delta(l_{i}, l_{j}) ] \mathcal{C}(\bm{{f}}^{(i)}, \bm{{f}}^{(j)}) + \epsilon}, \label{basic_loss}
\end{gather}
where $\mathcal{B}$ denotes the mini-batch, $\delta(\cdot, \cdot)$ is an indicator function and $\epsilon$ is a trivial number for computation stability.  
The naive design in (\ref{basic_loss}) is reasonable in theory and yet suffers from computational inefficiency. Since the complexity of the loss above is $\mathcal{O}(C_M^2)=\mathcal{O}(M^2)$, it increases quadratically as batch size $M$ goes bigger. Also the network  suffers from unstable parameter update and is hard to converge if we directly compute  loss from  two arbitrary samples from a mini-batch: a similar drawback as does in the triplet loss \cite{triplet_loss_xiaolong}.

%Inspired by the center loss \cite{center_loss}, 
As discussed previously, an effective solution to polymerize the inner-class feature distance is to provide a series of truncus in the network: serving as the centroid for each class and thus enforcing features to be learned around these hubs. To this end,
we define the \textit{centroid} of class $k$ as the average of features over a mini-batch: % $\mathcal{B}$:
%\begin{equation}
$
%\bm{c}_k = \frac{\sum_{i\in \mathcal{B}} \delta(l_{i}, k) \bm{{f}}^{(i)} }{ \sum_{i\in \mathcal{B}} \delta(l_{i}, k) + \epsilon} \in \mathbb{R}^{D  }. \label{centroid_def}
\bm{c}_k = \frac{1}{N_k}\sum_{i\in \mathcal{B}} \delta(l_{i}, k) \bm{{f}}^{(i)} \in \mathbb{R}^{D  }
$, where $N_k$ is the number of samples that belong to class $k$ within the batch.
%\end{equation}
%where $\epsilon$ is a trivial number for computation stability. 
Incorporating the spirit of class centroid into Eqn. (\ref{basic_loss}), one can \textcolor{black}{derive} the following revised loss to maximize:
\begin{equation}
%{p}_{l_i}^{(i)}
\mathcal{L}^{revise}
 = \sum_{i \in \mathcal{B}} \frac{ \exp \mathcal{C}(\bm{{f}}^{(i)}, \bm{c}_{l_i}) }
{ \sum_{m \neq l_i} \exp \mathcal{C}(\bm{{f}}^{(i)}, \bm{c}_m) },% \in \mathbb{R}. 
\label{coco_per_sample}
\end{equation}
where $m$ indexes along the category dimension. % in $K$.
The direct intuition behind Eqn. (\ref{coco_per_sample}) is to measure  the distance  of one sample against other samples by way of a class centroid, instead of a direct pairwise comparison as in Eqn. (\ref{basic_loss}). 
The numerator ensures  sample $i$ is close enough to its class center $l_i$ and the denominator enforces a minimal distance against samples in other classes. The exponential operator $\exp$ is to transfer the cosine similarity %$\mathcal{C} \in [-1, 1]$ 
to a normalized probability output. %, ranging from 0 to 1. 

To optimize the aforementioned loss in practice, we first normalize the feature and centroid by $l_2$ norm and then scale the feature before feeding them into the loss layer:
\begin{equation}
\hat{\bm{c}}_{k} =  \frac{\bm{c}_{k}} { \| \bm{c}_{k} \| },~\hat{\bm{f}}^{(i)} =  \frac {\alpha \bm{f}^{(i)}} {  \| \bm{f}^{(i)} \|},~p_{k}^{(i)} = \frac{\exp (    \hat{ \bm{c} }_{k}^{T} \cdot \hat{\bm{f}}^{(i)} ) }{  \sum_m  \exp (\hat{ \bm{c} }_{m}^{T} \cdot \hat{\bm{f}}^{(i)}  )}, \\
\end{equation}
where $\alpha$ is the scale factor and $p_{k}^{(i)}$ is an alternative expression\footnote{ In practice, it is more feasibly convenient to include $m=l_i$ in the denominator and the result barely differs.} to the inner term of summation in Eqn. (\ref{coco_per_sample}). 
Therefore, the proposed congenerous 
cosine (COCO) loss are formulated, in a cross-entropy manner to minimize, as follows:
%\begin{gather}
%\arg \min_{\bm{W}^*} \mathcal{L} = \arg \min_{\bm{W}^*} - \sum_{i \in \mathcal{B}} \log p_{l_i}^{(i)}. %+ \frac{\lambda}{2} \| \bm{W} \|^2.
%\end{gather}
\begin{gather}
\mathcal{L}^{COCO} \big(\bm{{f}}^{(i)}, \bm{c}_k \big) %=  \sum_{i \in \mathcal{B}}	\mathcal{L}^{(i)} 
= - \sum_{i \in \mathcal{B}, k}    t_k^{(i)} \log p_{k}^{(i)} = -\sum_{i \in \mathcal{B}} \log p_{l_i}^{(i)}, %+ \frac{\lambda}{2} \| \bm{W} \|^2.
\label{coco_loss}
\end{gather}
where $k$ indexes along the class dimension in $\mathbb{R}^K$ and $t_k^{(i)}\in \{0,1\}$ is the binary mapping of sample $i$ based on its label $l_i$.
The proposed COCO inherits the properties from (\ref{basic_loss}) to (\ref{coco_per_sample}), that is to increase the discrimination across categories as well as to polymerize the compactness within one class in a cooperative way. Moreover, it reduces the computation complexity compared with the naive version and can be implemented in a cheat way.
%$\lambda$ is a hyper-parameter of the regularization term.
% In practice, COCO loss can implemented in a neat way via the softmax operation.
%%
%For Eqn. \ref{coco_per_sample}, if we constrain the feature and centroid to be normalized 
%(\textit{i.e.}, $\hat{\bm{f}} = {\bm{f}} / { \| \bm{f} \| }$, $\hat{\bm{c}} = {\bm{c}}/ { \| \bm{c} \| }$) and loose the summation in the denominator to include $k=l_i$, the probability output of sample $i$ becomes:
%\begin{equation}
%p_{k}^{(i)} = \frac{\exp (    \hat{ \bm{c} }_{k}^{T} \cdot \hat{\bm{f}}^{(i)} ) }{  \sum_m  \exp (\hat{ \bm{c} }_{m}^{T} \cdot \hat{\bm{f}}^{(i)}  )}
%=\texttt{softmax} \big(
%%\hat{\bm{c}}_k, \hat{\bm{f}}^{(i)}
%z_k^{(i)}
%\big),
%\label{softmax_form}
%\end{equation}
%%
%where $z_k^{(i)}=\hat{ \bm{c} }_{k}^{T} \cdot \hat{\bm{f}}^{(i)}$ is the input to softmax. %The normalized centroids 
%$\hat{ \bm{c} }_{k}$ can be seen as  weights in the classification layer with  bias term being zero. \textcolor{black}{The advantage of COCO loss in Eqn. \ref{coco_loss} and \ref{softmax_form} from the naive version in Eqn. \ref{basic_loss} are two folds: it reduces the complexity of computation and could be achieved via the softmax with normalized inputs in terms of cosine distance.}

Note that both the features and  cluster centroids %$\bm{c}_k$ 
are %learnable parameters and 
trained end-to-end. 
We now derive the gradients of  loss $\mathcal{L}^{COCO}$ w.r.t. the unnormalized input %non-normalized 
feature $\bm{f}^{(i)}$ and the centroid $\bm{c}_k$. 
%written in an element-wise form (we drop the sample index $i$ for brevity), is derived as follows:
%\footnote{ We drop the sample index $i$ in $\bm{f}^{(i)}$ for brevity and use $d$ 
%%	(\textit{i.e.}, $f_d$) 
%	to index along the feature dimension in \bm{f}.}: 
For brevity, we drop the sample index $i$ and superscript notation in the loss. Denote 
$\nabla_{x} \mathcal{L}, x\in\{d,j\}$ as the element-wise top gradient of loss w.r.t. the normalized feature, \textit{i.e.}, 
$\partial \mathcal{L} / \partial \hat{f}_x $, we have the following gradient by applying the chain rule:
\begin{align}
%z_k & = \hat{ \bm{c} }_{k}^{T} \cdot \hat{\bm{f}},  \\ 
\frac{\partial \mathcal{L}}{ \partial f_j} & = \sum_d 
%\frac{\partial \mathcal{L}}{ \partial \hat{f}_d} 
\nabla_{d} \mathcal{L}
\cdot 
\frac{ \partial \hat{f}_d} { \partial {f}_j} %\nonumber \\
%
%
%& 
= \nabla_{j} \mathcal{L}
\cdot 
\frac{ \partial \hat{f}_j} { \partial {f}_j}    + \sum_{d \ne j}  
\nabla_{d} \mathcal{L}
\cdot 
\frac{ \partial \hat{f}_d} { \partial {f}_j}, \nonumber \\
%
%		& = \sum_k \frac{\partial \mathcal{L}^{(i)}}{ \partial z_k} \cdot \frac{ \partial z_k}{\partial \hat{f}_d} \cdot \frac{ \partial \hat{f}_d} { \partial {f}_d}, \nonumber \\
%
%		 	& = \sum_k (p_k - t_k) \cdot \hat{c}_{kd} \cdot \textcolor{black}{ \frac{ 1- \hat{f}_d^2  }{ \| \bm{f} \|}}, \\
& = \alpha \bigg( \frac {\nabla_{j} \mathcal{L}} {\| \bm{f} \| } - \sum_{d} \nabla_{d} \mathcal{L} \cdot \frac{f_i  f_d}{  \| \bm{f} \|^3  } \bigg)%, \nonumber\\
%
%
%	\frac{\partial \mathcal{L}^{(i)}}{ \partial c_{kd}} & = \frac{\partial \mathcal{L}^{(i)}}{ \partial \hat{c}_{kd}} \cdot  
%	\frac{ \partial \hat{c}_{kd}} { \partial c_{kd}}, \nonumber \\
%	& =  \frac{\partial \mathcal{L}^{(i)}}{ \partial z_k} \cdot \frac{ \partial z_k}{\partial \hat{c}_{kd}} \cdot \frac{ \partial \hat{c}_{kd}} { \partial c_{kd}}, \nonumber \\
%%	& =  (p_k - t_k) \cdot \hat{f}_{d} \cdot \frac{ 1- ( {\hat{\bm{c}}}^T_k \cdot \nabla \hat{\bm{c}}_k )\hat{c}_{kd}  }{ \| \bm{c}_k \|},
%%\textcolor{black}{
%	& =  (p_k - t_k) \cdot \hat{f}_{d} \cdot  \textcolor{black}{ \frac{ 1-\hat{c}_{kd}^2  }{ \| \bm{c}_k \|}}.
=  \frac { \alpha \big(
	\nabla_{j} \mathcal{L}        - (  \nabla \mathcal{L}^T \cdot \hat{\bm{f}}    )  \hat{f}_j       \big)
} {\| \bm{f} \| },
\end{align}
where $\nabla \mathcal{L} \in \mathbb{R}^D$ is the vector form of $\nabla_{x} \mathcal{L}$. Considering the specific loss defined in Eqn. (\ref{coco_loss}), the top gradient $\nabla_{x} \mathcal{L}$ can be obtained:
\begin{equation}
\nabla_{x} \mathcal{L}   
%\triangleq %\vcentcolon=
%\frac{\partial \mathcal{L}}{ \partial \hat{f}_x}
= \sum_k \frac{\partial \mathcal{L}}{ \partial z_k} \cdot \frac{ \partial z_k}{\partial \hat{f}_x}
= \sum_k (p_k - t_k) \cdot \hat{c}_{kx}.
\end{equation}
%$\hat{\bm{f}}$.  
The derivation of gradient w.r.t. centroid ${\partial \mathcal{L}}/{ \partial c_{kj}}$ can be derived in a similar manner.
The features are initialized from pretrain models and the initial value of $\bm{c}_k$ is thereby obtained.
% via Eqn.  (\ref{centroid_def}).

\subsection{Towards an Optimal Scale Factor $\alpha$}
As stated in previous subsection, we first normalize and scale the feature; now we have the following theorem to prove that there exists an optimal value for the factor. The derivations are provided in the supplementary.
\begin{theorem}
	Given the optimization loss $\mathcal{L}$ has an upper bound $\epsilon$, i.e, $\mathcal{L} < \epsilon$; and the neural network has a class number of $K$, the scale factor $\alpha$ enforced on the input feature has a lower boundary:
	\begin{equation}
	\alpha > \frac{1}{2} \log \frac{K-1}{ \exp \epsilon -1 }. \label{factor}
	\end{equation}\label{theorem}
\end{theorem}
\vspace{-.5cm}
%It is empirically found that a typical value for  $\epsilon$ is $10^{-4}$ and thus the scale factor can be expressed in a deterministic closed form:
%\begin{equation}
%\alpha = \frac{1}{2} \log(K-1) +3.
%\end{equation}

\begin{figure}
	\centering
	%\fbox{\rule{0pt}{1.5in} \rule{0.9\linewidth}{0pt}}
	\includegraphics[width=0.9\textwidth]{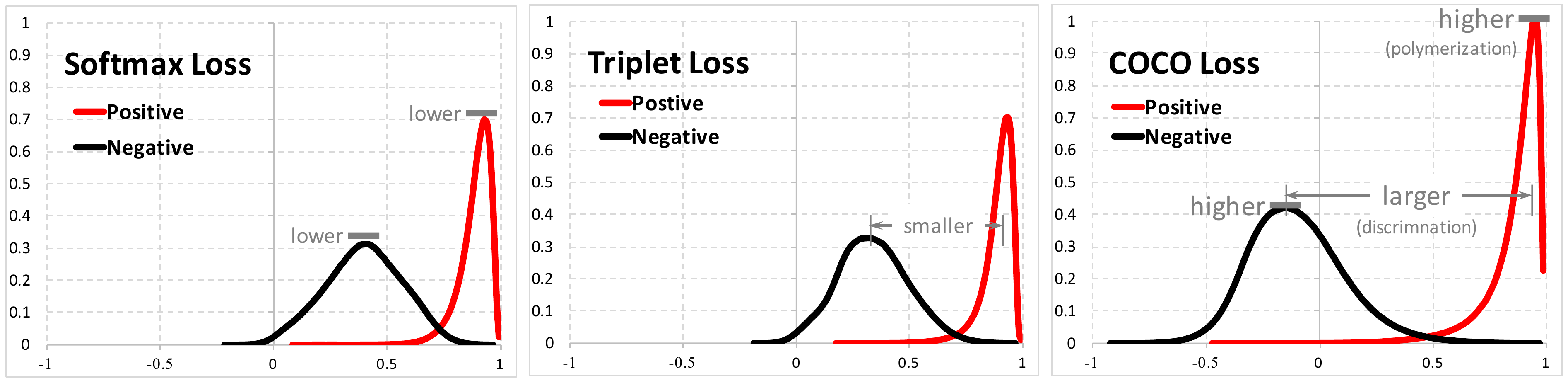}
%	\includegraphics[width=0.25\textwidth]{softmax}~~
%	\includegraphics[width=0.25\textwidth]{triplet}~~
%	\includegraphics[width=0.25\textwidth]{coco}
%	\vspace{-.1cm}
	\caption{Histogram statistics of cosine distance for positive and negative  pairs on MNIST \cite{mnist}. Feature discrimination and polymerization are more evident using COCO than other losses. }
	\label{fig:roc}
%	\vspace{-.7cm}
\end{figure}

\subsection{COCO Verification on Image Classification}
%\subsection{Relationship of COCO with  counterparts}
%\vspace{-.3cm}

In this subsection, we provide results and discussions to verify the effectiveness of COCO on small-scale image classification datasets, namely, MNIST \cite{mnist} and CIFAR-10 \cite{cifar}. The dataset, network structure and implementation details are provided in the supplementary.

\textbf{An optimal scale factor matters.} Take CIFAR-10 as example. When $\alpha$ is small ($\alpha=1$, error rate 12.4), the lower bound of loss is high, leading to a high loss in easy samples; however, the loss for hard examples still remains at low value. This will make the gradients from easy and hard samples stay at the same level. This is hard for the network to optimize towards hard samples - leading to a higher error rate. When   $\alpha=10$, the error rate is 7.22; if we set the factor to the optimal value defined in Eqn. (\ref{factor}), the error rate is 6.25.

\textbf{Feature discrimination and polymerization.} Figure \ref{fig:roc} shows the histogram of the cosine distance among positive pairs 
(\textit{i.e.}, two arbitrary samples that belong to the same class) %identity in \texttt{test\_0} and \texttt{test\_1}) 
and negative pairs.
%\fbox{\rule{0pt}{.7in} \rule{0.9\linewidth}{0pt}}
We can see that the distance in softmax and triplet cases resemble each other; in the COCO case, the discrepancy between the positive and negative is larger (center distance on \textit{x}-axis between two clusters); the intra-class similarity is also more polymerized (area under two clusters).

\begin{figure}[h]
	\centering
	\includegraphics[width=.85\textwidth]{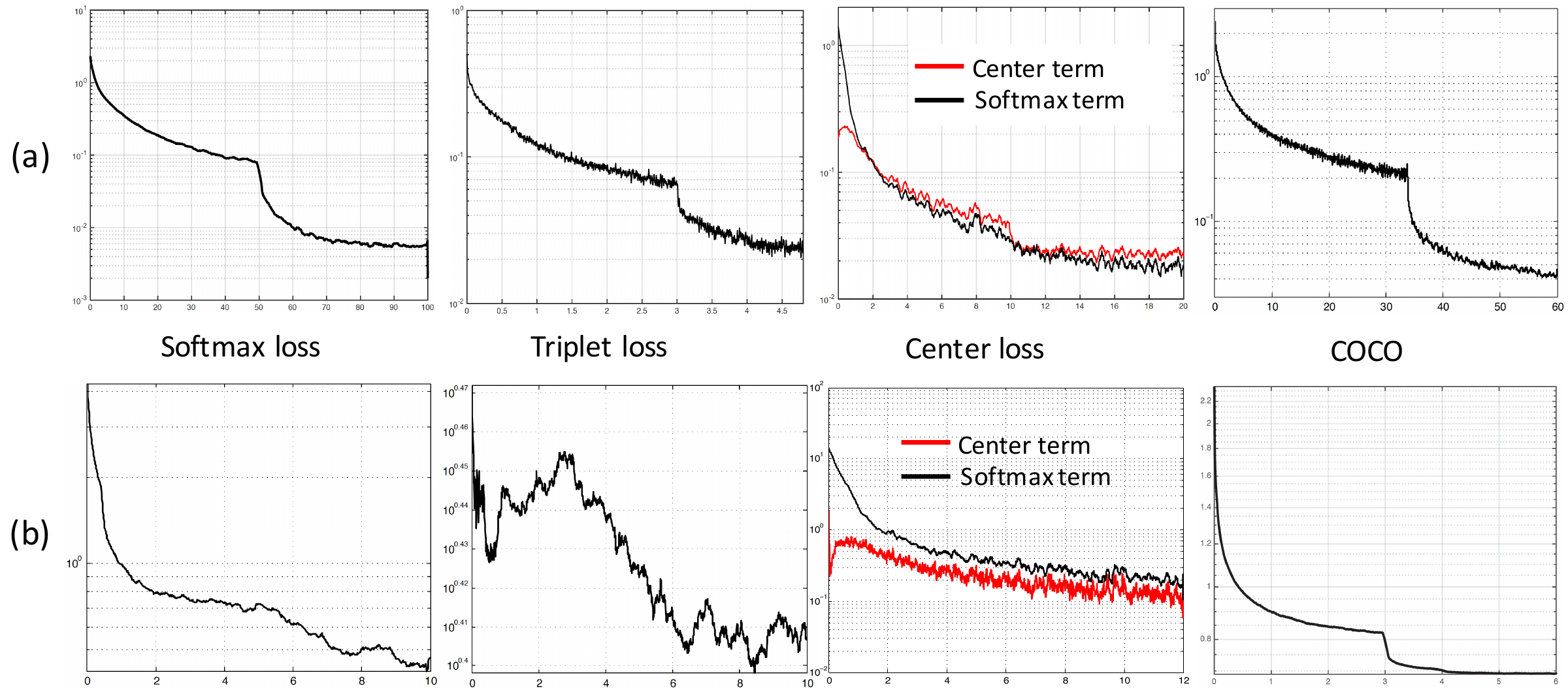}
	\caption{ Training loss vs epoch of different losses. (a): Image classification on CIFAR-10 \cite{cifar}; (b): Face verification on trained on MS-1M celebrity set. %LFW \cite{LFWTech}. 
		Note that for the center loss there are two curves, corresponding to the center and softmax term, respectively.
	}
	\label{fig:loss_temp}
\end{figure}

\textbf{Training loss.}
Figure \ref{fig:loss_temp} describes the training loss curves under different loss strategies on CIFAR and LFW datasets. We can observe that our proposed method undergoes a stable convergence as the epoch goes; whereas other losses do not witness an obvious drop when the learning rate decreases. 
The triplet loss is effective the number of classes is small on CIFAR; however, when the task extends to a larger scale and includes over 1,000 categories, the triplet term suffers from severe disturbance during training since the optimized feature distance is altered frequently within one batch.

\textbf{Quantitative results.}
We report the image classification error rate on MNIST and CIFAR-10 in Table \ref{tab:small_scale_cls}. 
%Our algorithm is superior to all previous state-of-the-arts. 
Note that MNIST is quite a smalls-scale benchmark and the performance tends to saturate. Moreover, we conduct the ablative study on different losses: softmax, center loss, triplet loss and their combination. For fair comparison, they all share the same CNN structure as COCO's. 
We can observe that the center loss alone bears a limited improvement (6.66 vs 6.70) over softmax; the triplet loss has the training instability concern and thus the error rate is much higher (12.69). Without resorting to softmax, our method is neat in formulation and achieves the lowest error (6.25).

\section{COCO for Large-scale  Recognition}

\subsection{Face Verification and Identification}\label{sec:face-verification-and-identification}
%\textcolor{blue}{TODO.}
We follow the general convention as in \cite{center_loss,facenet,sphereFace} to conduct face verification and identification: train a network using COCO loss to obtain robust features, extract features in the test set based on the trained model, and compare or identify the unknown classes given a large scale test setting. The main modifications to the traditional practice are two folds: first we employ the detection pipeline to leverage the feature selection and conduct landmark regression, in a multi-task RPN manner \cite{faster_rcnn}. One task is for classifying the foreground or background landmarks while the other focus on regressing them. There are five landmarks: left or right eyes, nose, and left or right corners of mouth. The second revision is the landmark alignment via affine transformation, in order to switch the position to a base one. % and prevent the model from over-fitting.

\begin{wraptable}{l}{7cm}
	\centering
	\captionof{table}{Classification error rate (\%) on MNIST and CIFAR-10 without data augmentation.
	} \label{tab:small_scale_cls}
	%		\smallskip
	\vspace{-.3cm}
	\footnotesize{			
		\begin{tabular}{l c c}
			\toprule
			Method                 & MNIST & CIFAR-10 \\
			\midrule
			DropConnect \cite{drop_connect}            & 0.57  & 9.41     \\
			NIN \cite{NIN}                    & 0.47  & 10.47    \\
			Maxout  \cite{maxout}               & 0.45  & 11.68    \\
			DSN  \cite{DSN}                  & 0.39  & 9.69     \\
			R-CNN  \cite{rcnn}                & 0.31  & 8.69     \\
			GenPool \cite{gen_pool}               & 0.31  & 7.62     \\
			\midrule
			Softmax           & 0.36  & 6.70      \\
			Center loss + softmax  & 0.32 
			& 6.66     \\
			Triplet loss           & 1.45  & 12.69    \\
			Triplet loss + softmax & 0.38  & 6.73     \\
			\midrule
			COCO                   & \textbf{0.30}   & \textbf{6.25}    \\
			\bottomrule
		\end{tabular}
	}	
\end{wraptable}

\subsection{Person Recognition}\label{sec:person-recognition}

Following a general pipeline of the verification task, we first train features using the proposed COCO algorithm on a specific body region (\textit{e.g.}, head) to classify different identities; during inference, the features on both test subsets are extracted to compute the cosine distance and to determine whether both instances are of the same identity. We utilize the multi-region 
%parallel processing 
spirit \cite{person_recog_rnn} during training and test to merge results from four regions, namely, \textit{face, head, upper body} and \textit{whole body}.
The contributions as regard to this task are two folds: remove the second training on \texttt{test\_0} as did in previous work, which credits from the property of COCO to learn discriminative and polymerized features; align each region patch to a base location to reduce variation among samples.

The detailed description of region detection and patch alignment are provided in the supplementary.
Given the cropped and aligned patches, we train (finetune) models for different regions on PIPA \texttt{training} set using COCO loss. 
%
%\textbf{Inference.} 
At testing stage, we measure the cosine similarity between two test splits to recognize the identity in \texttt{test\_1} based on the labels in \texttt{test\_0}.
The similarity between two patches $i$ and $j$ in \texttt{test\_1} and \texttt{test\_0} is denoted as
$s(r)_{ij}$, where $r$ indicates a  region. 
%A key problem is how to merge the similarity scores from different regions. 
We %first 
normalize the preliminary  $s(r)_{ij}$ in order to have scores across different regions comparable:
%, which is achieved by the logistic regression; 
\begin{gather}
	s(r)_{ij} = \mathcal{C}(\bm{f}^{(i, r)}, \bm{f}^{(j, r)}),~~~\hat{s}(r)_{ij}  = \bigg( 1 + \exp \big[- \big(  \beta_0 + \beta_1 s(r)_{ij}  \big)  \big]  \bigg)^{-1},
\end{gather}
where $\beta_0, \beta_1$ are parameters of the logistic regression.
The final score $S_{ij}$ is a weighted mean of the normalized scores $\hat{s}(r)_{ij} $ of each region:
$S_{ij} = \sum_{r=1}^{R} \gamma^r \cdot \hat{s}(r)_{ij}$, where
$R$ is the total number of regions and $\gamma^r $ being the weight of each region's score.
The identity of patch $i$  in \texttt{test\_1} is decided by the label corresponding to the maximum score in the reference set:
$l_i = \arg \max_{j*} S_{ij}$.
%
% 
%Such a scheme guarantees that when new training data are added into \texttt{test\_0}, there is no need to train a second model or SVM on the reference set, which is quite distinct from previous work. 
\textcolor{black}{The test parameters of $\beta$ and $\gamma$ are determined by a validation set on PIPA.}

%\vspace{-.5cm}
\section{Experiments}

%\subsection{Setup}
\textbf{Dataset and evaluation metric.
}
\textit{Face recognition.}
%\textcolor{blue}{TODO.}
The Labeled Face in the Wild (LFW) dataset \cite{LFWTech} contains 13,233 web-collected images from 5749 different identities,
with large variations in pose, expression and illuminations. Following
the standard protocol of unrestricted with labeled outside data, we test on
6,000 face pairs.
The MegaFace \cite{kemelmacher2016megaface} is a very
challenging dataset and aims to evaluate the performance of face recognition
algorithms at the million scale of distractors: people who are not in the
test set. It includes gallery set and probe set. The gallery
set consists of more than one million images from 690K different individuals, as a
subset of Flickr photos from Yahoo. The probe set descends from two existing databases: Facescrub and FGNet.
\textit{Person recognition.} The People In Photo Albums (PIPA) dataset \cite{piper} is divided into train, validation, test and leftover sets, where the head of each instance is annotated in all sets and it consists of over 60,000 instances of around 2,000 individuals.
%As did in \cite{oh_iccv,person_recog_rnn,piper,zlin}, the training set
%is only used for learning feature representations;
%%
%the recognition system is trained on \texttt{test\_0} and evaluated on \texttt{test\_1}. 
In this work, thanks to the discriminative and polymerized features learned by COCO,
we take full advantage of the training set and remove the second training on \texttt{test\_0}.
Moreover, \cite{oh_iccv} introduced three more challenging splits besides the \texttt{original} test split, 
namely \texttt{album}, \texttt{time} and \texttt{day}. Each new split emphasizes different temporal distance (various albums, events, days, etc.)
between the two subsets of the test data.
The evaluation metric is the averaged classification accuracy over all instances on \texttt{test\_1}.

\subsection{Face Verification and Identification}

\begin{figure}
	\begin{floatrow}
		\ffigbox{%
				\includegraphics[width=.45\textwidth]{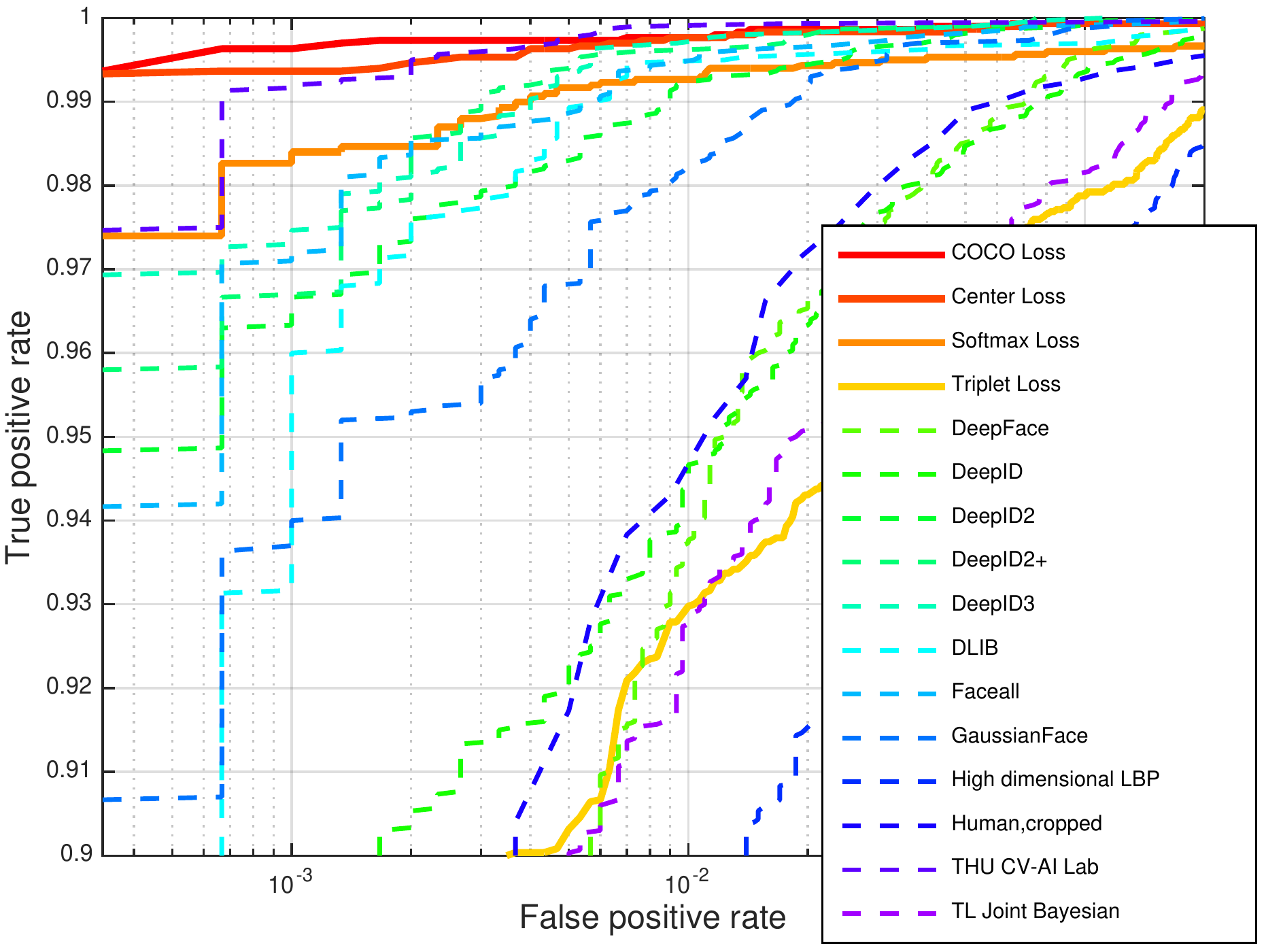}
		}{%
		\caption{The ROC curves of different methods for face verification on LFW. 
			%Results of other approaches are directly from the dataset website. 
			Our model using COCO to train features achieves a higher AUC score.}\label{roc}%
	}
	\capbtabbox{%
		\footnotesize{
			\begin{tabular}{l c c }
				\toprule
				Method                     & Train Data    & mAcc   \\
				\midrule
				FaceNet   \cite{facenet}                   & 200M    & 99.65       \\
				DeepID2 \cite{contrastive_loss}   *                  & 300K     & 99.47        \\
				CenterFace \cite{center_loss}                  & 700K    & 99.28        \\
				L-Softmax  \cite{quite_similar_loss_to_coco}                  & \texttt{webface} & 98.71        \\
				SphereFace \cite{sphereFace}                  & \texttt{webface} & 99.42       \\
				\midrule
				Softmax                      & \texttt{half MS-1M} & 99.75   \\
				Center loss + softmax        & \texttt{half MS-1M} & 99.78  \\
				Triplet loss                 & \texttt{half MS-1M} & 98.85  \\
				Triplet loss + softmax       & \texttt{half MS-1M} & 99.68 \\
				\midrule
				COCO                         & \texttt{half MS-1M} & \textbf{99.86}  \\
				\bottomrule
			\end{tabular}
		}
	}{%
	\caption{Face verification accuracy (\%) on LFW. The first three models use private outside training data.
		% while the rest %(\texttt{webface})
		%utilizes the publicly available CAISA-webface 
		%data. 
		* Result %*in DeepID2 \cite{contrastive_loss} 
		comes from an ensemble method that has 25 models.
	} \label{tab:lfw}
}
\end{floatrow}
\vspace{-.5cm}
\end{figure}

%\textbf{Comparison to state-of-the-arts.}
We now apply the COCO feature learning to large-scale human recognition system. For fair comparison of our method to other losses (softmax, center loss, triplet loss and their combination), we use the same CNN network structure in the following experiments. Features on LFW is available on the github repository.

Figure \ref{roc} shows the ROC curves of different methods and Table \ref{tab:lfw} reports the accuracy (\%) for face verification on LFW \cite{LFWTech}.
It is observed that COCO model is competitively superior (99.86) than its counterparts. Compared with FaceNet \cite{facenet} which intakes around 200M outside training data, or DeepID2 \cite{contrastive_loss} which ensembles 25 models, our approach demonstrates the effectiveness of feature learning.
% given a limited training data. 
Table \ref{tab:megaface} illustrates the identification accuracy on MegaFace challenge \cite{kemelmacher2016megaface}, where the test identities share no interaction with those in training and the data scale is quite large. Our model using COCO is competent at different number of distractors. The combination of center loss with softmax ranks second (76.57 vs 75.79 @1M), which verifies that applying center loss alone cannot guarantee a good feature learning for large-scale task. The triplet loss (69.13) is inferior and our model competes favorably against those using  \texttt{large} protocol of training data.

\begin{table}
	\begin{center}
		\footnotesize{
			\begin{tabular}{l c c c c c}
				\toprule
			Method                 & Protocol & @1M   & \multicolumn{1}{l}{Method}         & Protocol & @1M   \\
			\midrule
			NTechLab               & \texttt{small}    & 58.22 & \multicolumn{1}{l}{NTechLab}       & \texttt{large}    & 73.30  \\
			DeepSense              & \texttt{small}    & 70.98 & \multicolumn{1}{l}{DeepSense}      & \texttt{large}    & 74.80  \\
			SphereFace \cite{sphereFace}            & \texttt{small}    & 75.77 & \multicolumn{1}{l}{ShanghaiTech}   & \texttt{large}    & 74.05 \\
			L-Softmax \cite{quite_similar_loss_to_coco}             & \texttt{small}    & 67.13 & \multicolumn{1}{l}{Google FaceNet} & \texttt{large}    & 70.50  \\
			\midrule
			\midrule
			Method                 & Protocol & @1M   & @100k          & @10k     & @1k   \\
			\midrule
			Softmax                & \texttt{small}    & 71.17 & 85.78          & 93.22    & 96.14 \\
			Center loss + softmax  & \texttt{small}    & 75.79 & 90.74          & 96.45    & 97.91 \\
			Triplet loss           & \texttt{small}    & 69.13 & 85.64          & 93.38    & 96.66 \\
			Triplet loss + softmax & \texttt{small}    & 70.22 & 86.77          & 94.71    & 97.02 \\
			\midrule
			COCO                   & \texttt{small}    & \textbf{76.57} & \textbf{91.77}          & \textbf{96.72}    & \textbf{98.03} \\
				\bottomrule
				
			\end{tabular}
			
		}
		\vspace{-.3cm}
		\caption{Face identification accuracy (\%) on MegaFace challenge at various number of distractors.
		} \label{tab:megaface}
		
	\end{center}
	
\end{table}

%\textbf{Ablation study on scale factor. }
%\textcolor{blue}{TODO.}

\subsection{Person Recognition}

%The comparison with state-of-the-arts and visual results for person recognition is provided in the supplementary material due to page limit. 
%\begin{wraptable}{l}{8.5cm}
%	\begin{center}
%		\vspace{-.4cm}
%		\footnotesize{
%			\begin{tabular}{c c c c c}
%				\toprule
%				Methods & \texttt{ original} & \texttt{ album} & \texttt{ time}  & \texttt{ day}
%				\\
%				\midrule
%				PIPER
%				\cite{piper} 
%				& 83.05 & - & - & -\\
%				RNN
%				\cite{person_recog_rnn} 
%				& 84.93 & 78.25 & 66.43 & 43.73 \\
%				Naeil
%				\cite{oh_iccv} 
%				& 86.78 & 78.72 & 69.29 & 46.61 \\
%				%				Li \textit{et al.} 
%				Multi-context
%				\cite{zlin} & 88.75 & 83.33 & 77.00 & 59.35 \\
%				\midrule
%				Ours & \textbf{92.78} & \textbf{83.53 }& \textbf{77.68 }& \textbf{61.73 }\\
%				\bottomrule
%			\end{tabular}
%		}
%		\vspace{-.3cm}
%		\caption{Verification accuracy (\%) comparison to state-of-the-arts on PIPA \cite{piper} for person recognition.
%		} \label{tab:compare}
%		%		\smallskip
%	\end{center}
%\end{wraptable}
\textbf{Comparison with state-of-the-arts and visual results.} The comparison figure and visual results are provided in the supplementary due to page limit. 
We can see %from Table \ref{tab:compare} 
that our recognition system outperforms against previous state-of-the-arts %, PIPER \cite{piper}, RNN \cite{person_recog_rnn}, Naeil \cite{oh_iccv}, 
in all four test splits. Note that the test identities share no interaction with those in the training set and thus this task is also a challenging open set verification problem.
%
%We also %Figure \ref{fig:visualize} 
The visual results show some examples of the predicted instances by our model, % in the supplementary, 
where complex scenes with non-frontal faces and body occlusion can be handled properly in most scenarios. 
The last two columns show 
failure cases where our  method could not differentiate the identities, 
this is probably due to the very similar appearance configuration in these scenarios.
% (same-view of frontal face, similar clothes and background).	

\begin{table*}[t]
	\begin{center}
		\caption{Person recognition on PIPA \cite{piper}. Investigation on the merging  similarity score from different body regions during inference. The top two results in each split set are marked in  {\textbf{bold}} and {}{\textit{\underline{italic}}}, respectively. COCO loss is applied in all cases if not specified.}  \label{tab:score_merge}
		%		\smallskip
%		\vspace{.2cm}
		%		\small{
		\footnotesize{
			\begin{tabular}{c | c c c c | c c c c}
				\toprule
				% after \\: \hline or \cline{col1-col2} \cline{col3-col4} ...
				Method & Face  & Head &  Upper body& Whole body & \texttt{original} & \texttt{album} & \texttt{time} &\texttt{day}
				\\
				\midrule
				-& \checkmark & \checkmark & && 84.17 & 80.78 & 74.00 & {}{{53.75}}\\
				-& \checkmark & & \checkmark && 89.24 & 81.46 & \underline{\textit{76.84}} & {}{\textit{\underline{61.48}}}\\
				-& \checkmark && & \checkmark & 88.40 & 82.15 & 70.90 & 57.87\\
				-& & \checkmark &\checkmark  && 88.76 & 79.15 & 68.64 & 42.91\\
				RNN \cite{person_recog_rnn} & & \checkmark &\checkmark  && 84.93 & 78.25 & 66.43 & 43.73\\
				-& & \checkmark & &\checkmark & 87.43 & 77.54 & 67.40 & 42.30\\
				-& & & \checkmark & \checkmark & 81.93 & 73.84 & 62.46 & 34.77 \\
				\midrule
				-& \checkmark & \checkmark & \checkmark&& 87.86 & 80.85 & {}{{{71.65}}} & 59.03 \\
				-& \checkmark & \checkmark & &\checkmark & {}{{{88.13}}} &{}{\textit{\underline{82.87}}} &73.01 &55.52\\
				-& & \checkmark & 	\checkmark &	\checkmark & 89.71 & 78.29 & 66.60 & 52.21 \\
				-& \checkmark & &\checkmark  &\checkmark & \underline{\textit{91.43 }}& 80.67 & 70.46 & 55.56  \\
				\midrule
				Softmax& \checkmark & \checkmark & \checkmark &\checkmark & {}{{88.73}} & {}{{80.26}} & {}{{71.56}} & 50.36 \\
				-& \checkmark & \checkmark & \checkmark &\checkmark & {}{\textbf{92.78}} & {}{\textbf{83.53}} & {}{\textbf{77.68}} & \textbf{61.73} \\
				\bottomrule
			\end{tabular}
		}
	\end{center}
\end{table*}

\textbf{Ablation study on different regions.} Table \ref{tab:score_merge} depicts the investigation on merging the similarity score  from different body regions during inference. %at test time. 
%Generally speaking, taking all  regions into consideration could result in the best accuracy of 92.78 on the \texttt{original}  set. 
%%
%It is observed that the performance is still fairly good on  \texttt{day}  and  \texttt{time}  if the two scores of face and upper body alone are merged. 
%
%It is also proved that the face region is of vital importance as removing it could achieve lower accuracies (81.47 and 86.42 on \texttt{original}, for example).
%
\cite{person_recog_rnn} also employs a  multi-region processing step and we include it in Table  \ref{tab:score_merge}.
For some region combination,
on certain splits our model is superior (\textit{e.g.}, head plus upper body region, 88 vs 84 on \texttt{original}, 79 vs 78 on \texttt{album}, 68 vs 66 on \texttt{time}); whereas on other splits ours is inferior (\textit{e.g.}, 42 vs 43 on \texttt{day}). This is probably due to distribution imbalance among %different 
splits: the upper body region differs greatly in appearance with some instances absent of this region, making COCO  hard to learn the corresponding features. However, under the score integration scheme, the final prediction (last line) can complement features learned among different regions and achieves better performance against \cite{person_recog_rnn}.

\section{Conclusion}

In this work, we re-investigate the feature learning problem based on the motivation that deeply learned features should be both discriminative and polymerized. We address the problem by proposing a congenerous cosine (COCO) loss, which optimizes the cosine distance among data features to simultaneously enlarge inter-class variation and intra-class similarity. COCO can be learned in a neat way with stable end-to-end training. We have extensively conducted experiments on five benchmarks, from image classification to face identification, to demonstrate the effectiveness of our approach, especially applying it to the large-scale human recognition tasks.

\bibliographystyle{ieee.bst}
\bibliography{deep_learning,icme2017template}
\clearpage

\section*{Supplementary}
\subsection{Proof on the optimal scale factor, Theorem \ref{theorem}}

%\begin{theorem}
%	Given the optimization loss $\mathcal{L}$ has an upper bound $\epsilon$, i.e, $\mathcal{L} < \epsilon$; and the neural network has a class number of $K$, the scale factor $\alpha$ enforced on the input feature has a lower boundary:
%	\begin{equation}
%	\alpha > \frac{1}{2} \log \frac{K-1}{ \exp \epsilon -1 }.
%	\end{equation}
%\end{theorem}

\begin{proof}
	It is obvious that the infimum value of COCO loss is of the form: 
	\begin{gather}
	\inf \mathcal{L}^{COCO} = - \log \frac{\exp s^{+}}{ \exp s^{+} + (K-1) \exp s^{-} }, \\
	s^{+} = \alpha \exp (\cos \theta) \leq \alpha,\\
	s^{-} = \alpha \exp (\cos \theta) \geq -\alpha,
	\end{gather}
	where $s$ is the probability output in the final classification layer. For brevity, we denote $\inf s = s^{-}$ and  $\sup s = s^{+}$; $\theta$ is the angle between two features.
	Often we have $\mathbb{E}[\cos \theta] \geq 0$ in practice and yet the minimal value of $\cos \theta$ can still be $-1$. Therefore,
	\begin{align}
	%\inf
	\sup 
	\mathcal{L}^{COCO} =& - \log \frac{\exp (-\alpha) }{ \exp (-\alpha) + (K-1) \exp  \alpha }. %, \\
%	=& - \log \frac{1}{K} = \log K. \nonumber
	\end{align}
	The loss to minimize in an optimization problem usually meets some upper bound criteria, and we denote this target value as $\epsilon$, \textit{i.e.},
	$\inf \mathcal{L}^{COCO} < \epsilon$. Incorporating this prior knowledge into previous derivations, we have:
	\begin{gather}
	\log( \exp \alpha + (K-1) \exp (- \alpha)  ) < \alpha + \epsilon, \\
	\alpha > \frac{1}{2} \log \frac{K-1}{ \exp \epsilon -1 }.
	\end{gather}
\end{proof}

It is empirically found that a typical value for  $\epsilon$ is $10^{-4}$ and thus the scale factor can be expressed in a deterministic closed form:
\begin{equation}
\alpha = \frac{1}{2} \log(K-1) +3.
\end{equation}

\subsection{Details on Image Classification and Face Recognition}
For image classification on CIFAR, we use a simple ResNet-20 structure and train the network from scratch. We set the momentum as 0.9 and the
weight decay to be 0.005. The base learning rate is set to
be 0.1, 0.1, 0.05, respectively. We drop the learning rate by
10% around every 40 epoches in a continuous exponential
way and stop to decrease the learning rate until it reaches a
minimum value (0.0001). All the convolutional layers are initialized with
Gaussian distribution with mean of zero and standard variation
of 0.05 or 0.1.

For face recognition, we use the Inception ResNet model %\cite{resNet} 
and add a new fully-connected layer after the final global pooling layer to generate a 128-dim feature. 
The model is trained from scratch. The training data is a subset of the Microsoft-1M celebrity dataset, which consists of 80k identities with 3M images in total, \textbf{removing all overlapping IDs}. For COCO, softmax and center loss, we train the network with SGD solver with a base learning rate 0.01 and drop it to 10\% every five epoch. The total training time is 12 epoch. The momentum is set to 0.9. Batch normalization \cite{bn} with moving average is employed. 
%
%The batch size is 128.
%
%
For the 
triplet loss, we have a base learning rate to 0.1 with a bigger batch size 128*3 (it is hard to converge if the size is less than 64*3); we  drop the rate to 10\% every 90k iteration. The total training time is around 300k iterations.

\textbf{Evaluation metric on face recognition.}
%We evaluate coco loss on large scale face verification and identification tasks.
For verification, the test set is LFW \cite{LFWTech} %, 13,233 images of 5171 people
and the 
metric is under standard `Unrestricted, Labeled Outside Data' setting of LFW. It verifies 6,000 pairs to determine if they belong to the same person. We report ROC curve and the mean accuracy.
For identification on MegaFace \cite{kemelmacher2016megaface},
we have the test set into two subsets: the first is probe set that consists of FaceScrub (3530 images of 80 persons); the second is gallery set that consists of distractors (1M images with no overlap with FaceScrub).
The test metric is under standard MegaFace Challenge-1 setting. We report the CMC curve in terms of Top-1 accuracy with 10, 100, 1000, 10k, 100k and 1M distractors. 
%; in the paper, it is of a table form).

\subsection{More Algorithm Details for Person Recognition}

%\textbf{Region detection and training.} 
The COCO features are trained on four regions (thus four models), namely, \textit{face, head, upper body} and \textit{whole body}. We take the face region as an illustration for the alignment scheme. A face detector is first pretrained in a RPN spirit \cite{faster_rcnn}.
The face detector identifies some keypoints (eye, brow, mouth, \textit{etc}.) and we align the detected face patch to a base location via an affine transformation. 
Let  $p, q \in \mathbb{R}^{z \times 2} $ denote  $m$ keypoints and the aligned locations, respectively. We define $P,Q$ as two affine spaces, then the transformation
$\mathcal{A}: P \mapsto Q$ is defined as: 
%\begin{equation}
$p \mapsto q = \bm{A} p +b$, 
%\end{equation}
where $\bm{A }\in  \mathbb{R}^{z \times z} $ is a linear transformation matrix in $P$ and $b  \in  \mathbb{R}^{z \times 2} $ being the bias in $Q$.
%
%Note that %in some cases 
%in practice,
%if the confidence of a keypoint is below some threshold, we do not depend on such a point to align the patch; when the number of keypoints is less than three, we heuristically obtain the aligned patch based on the ground truth of the head, since at least three points can determine an affine transformation.
\textcolor{black}{
	%Such an alignment scheme is to ensure samples within each  category do not have large variance %with each other 
	%and to prevent the model from overfitting.
	Such an alignment scheme ensures samples both within and across categories do not have large variance: if the model is learned without alignment, it has to distinguish more patterns, \textit{e.g.}, different rotations among persons, making it more prone to overfitting; %if the network is equipped 
	with alignment, it can better classify 
	features of various identities despite of rotation, viewpoint, translation, \textit{etc}.
}
%\subsection{Robust feature representation}
Given the cropped and aligned  patches, we finetune the face model on PIPA \texttt{training} set using COCO loss. 

\begin{figure}
	\centering
	%\fbox{\rule{0pt}{2in} \rule{0.9\linewidth}{0pt}}
	\includegraphics[width=.98\textwidth]{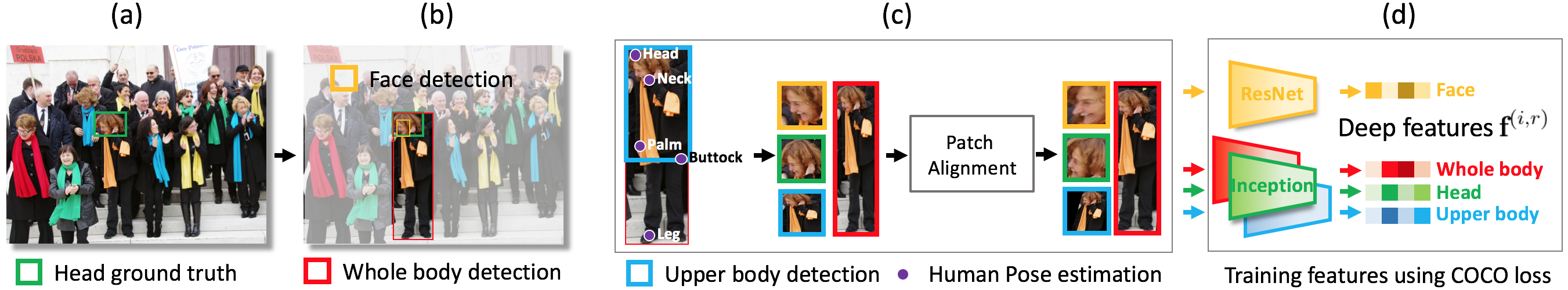}
	\vspace{.3cm}
	\caption{Training workflow of person recognition task. 
		(a) Each person is labelled with a ground-truth box of the head. 
		%		(b) For a typical training sample, we first detect %the 
		%		face and body regions. (c) Pose estimator \cite{pose} is employed to identify keypoints of a human body in order to find the upper body region. 
		(b): Face, whole body and upper body detection.
		(c): Each region (patch) is aligned to a base position to alleviate the inner-class variance. (d) Each aligned patch is further fed into a deep model to obtain representative and robust features using COCO loss.
		% is applied afterwards.
	}
	\label{fig:pipeline}
\end{figure}

The head region is given as the ground truth for each person.
%and the detection of face is stated previously. 
To detect a whole body, we also pre-train a detector in the RPN framework.
The model is trained on the large-scale human celebrity dataset \cite{celebrity},
where we use the
first 87021 identities in 4638717  images.
%for training verification network.
The network structure is an inception model \cite{bn} with the final pooling layer replaced by a fully connected layer.
%For 
To determine the upper body region, we  conduct human pose estimation \cite{pose}
to identity keypoints of the body and the upper part is thereby located by these points.
The head, whole body and upper body models, which are used for COCO loss training, are finetuned on PIPA training set using the pretained inception model, following similar procedure of patch alignment stated previously for the face region.
%in Section \ref{sec:face-detection-and-alignment}. 
The aligned patches of four regions are shown in Figure \ref{fig:pipeline}(c).

%\textcolor{blue}{List the model table here.}

\subsection{Experiments on Person Recognition}

% for person/face recognition
\begin{table*}
	\begin{center}
		\caption{Ablation study on body regions and feature alignment using softmax as loss. We report the classification accuracy 
			(\%) 
			%of the  {non-alignment} and {alignment} cases in left and right column within each region category, respectively. 
			where \texttt{n-a} denotes the non-alignment case and \texttt{ali} indicates the alignment case.
			Note that for  face region, we only evaluate instances with faces.}
		\label{tab:region}
		%		\smallskip
		\vspace{.2cm}
		\footnotesize{
			\begin{tabular}{c | c c | c c c | c c c | c c }
				\toprule
				\multirow{2}{*}{
					Test split } & \multicolumn{2}{c|}{Face} & \multicolumn{3}{c|}{Head} 
				& \multicolumn{3}{c|}{Upper body} & \multicolumn{2}{c}{Whole body}   \\
				& \texttt{n-a} & \texttt{ali} & \texttt{n-a} & \texttt{ali}& \cite{person_recog_rnn} & \texttt{n-a} & \texttt{ali}& \cite{person_recog_rnn} & \texttt{n-a} & \texttt{ali} \\
				\midrule 
				\texttt{ original} &  95.47 & 97.45 & 74.23 & 82.69  & 81.75 & 76.67 & 80.75  & 79.92 & 75.04 & 79.06 \\
				\texttt{ album}    & 94.66 & 96.57 & 65.47 & 73.77 & 74.21& 66.23 & 69.58 & 70.78 & 64.21 & 67.27  \\
				\texttt{ time}  & 91.03 & 93.36 & 55.88  & 64.31 & 63.73& 55.24  &  57.40  & 58.80& 55.53 & 54.62   \\
				\texttt{ day} & 90.36 & 91.32 & 35.27 & 44.24 & 42.75 & 26.49  & 32.09 &34.61& 32.85 & 29.59   \\
				\bottomrule
			\end{tabular}
		}
	\end{center}
	% 	\vspace{-.5cm}
\end{table*}
\textbf{Ablation study on feature alignment for different regions.} Table \ref{tab:region} reports the performance of using feature alignment and different
body regions, where several remarks could be observed. First, the alignment case in each region performs better by a large margin than the non-alignment case, which verifies the motivation of patch alignment to alleviate inner-class variance stated in 
%Section \ref{sec:face-detection-and-alignment}
the main paper. Second, for the alignment case, the most representative features to identify a person reside in the region of face, followed by head, upper body and whole body at last.
Such a clue is not that obvious  for the non-alignment case. Third, we notice that for the whole body region,  accuracy in the non-alignment case is higher than that of the alignment case in  \texttt{time} and \texttt{day}. This is probably due to the improper definition of base  points on these two sets.
%TODO: compare with RNN method

\begin{table}
	\begin{center}
		%\vspace{-.4cm}
		\caption{Verification accuracy (\%) comparison to state-of-the-arts on PIPA \cite{piper} for person recognition.
		} \label{tab:compare}
		\smallskip
		\footnotesize{
			\begin{tabular}{c c c c c}
				\toprule
				Methods & \texttt{ original} & \texttt{ album} & \texttt{ time}  & \texttt{ day}
				\\
				\midrule
				PIPER
				\cite{piper} 
				& 83.05 & - & - & -\\
				RNN
				\cite{person_recog_rnn} 
				& 84.93 & 78.25 & 66.43 & 43.73 \\
				Naeil
				\cite{oh_iccv} 
				& 86.78 & 78.72 & 69.29 & 46.61 \\
				%				Li \textit{et al.} 
				Multi-context
				\cite{zlin} & 88.75 & 83.33 & 77.00 & 59.35 \\
				\midrule
				Ours & \textbf{92.78} & \textbf{83.53 }& \textbf{77.68 }& \textbf{61.73 }\\
				\bottomrule
			\end{tabular}
		}
		%\vspace{-.3cm}
		
		%		\smallskip
	\end{center}
\end{table}

\begin{figure*}[h]
	\centering
	%\fbox{\rule{0pt}{2in} \rule{0.9\linewidth}{0pt}}
	\includegraphics[width=.95\textwidth]{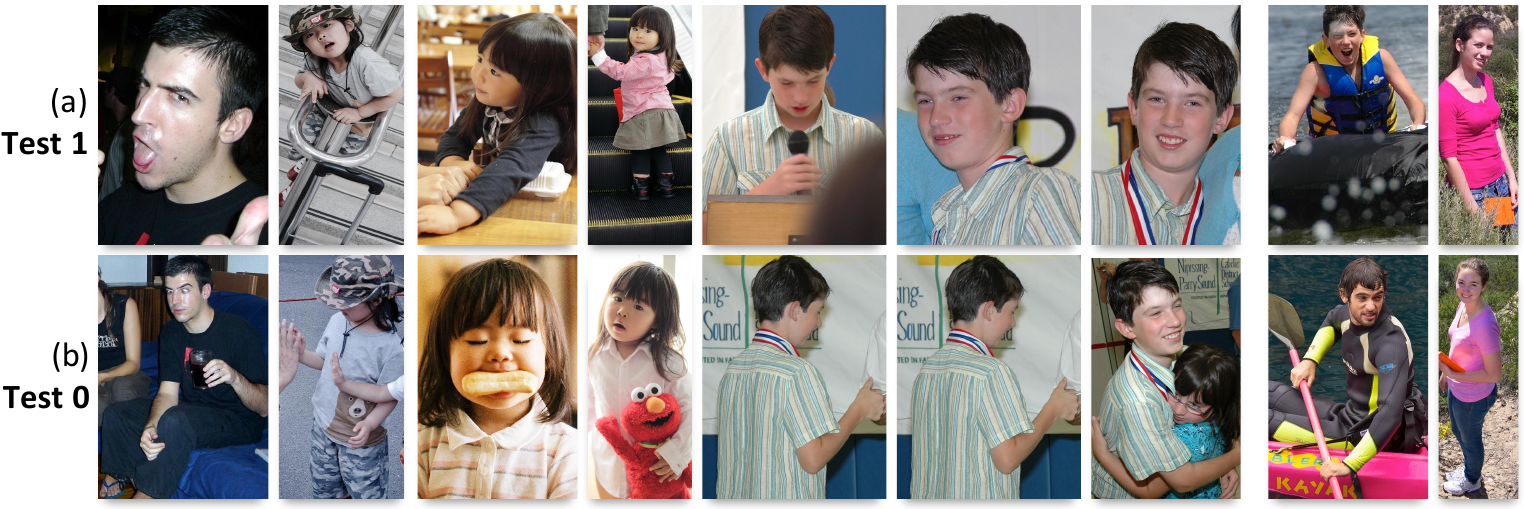}
	\vspace{.3cm}
	\caption{Visualization of our method on the PIPA \texttt{test\_1} set. 
		Given the %instance patch cropped from the 
		image as input to be predicted in (a), its nearest neighbour in the feature space from  \texttt{text\_0} is shown in (b). The identity of the input is determined by the label of its neighbour.
		Our model can handle complex scenes with non-frontal faces and body occlusion. 
		The last two columns show failure cases.
	}
	\label{fig:visualize}
\end{figure*}

\textbf{Visual results.} Figure \ref{fig:visualize} shows some visualization results for person recognition task.

\end{document}